\documentclass[letterpaper, 10 pt, conference]{ieeeconf}  

\IEEEoverridecommandlockouts                              

\usepackage{graphicx}
\usepackage{subfig}
\usepackage{color, colortbl}
\usepackage{booktabs}
\usepackage{multirow}
\usepackage{hyperref}
\usepackage[normalem]{ulem}

\usepackage{setspace}

\usepackage{amsmath} 
\usepackage{amssymb}  
\usepackage{hhline}
\usepackage{cleveref}
\usepackage[noadjust]{cite}

\begin{document}

\title{\LARGE \bf
EVOLIN Benchmark: Evaluation of Line Detection and Association 
}

\author{Kirill Ivanov$^{1, 2}$ \quad Gonzalo Ferrer$^{1}$ \quad Anastasiia Kornilova$^{1}$
\thanks{*This work was not supported by any organization}
\thanks{$^{1}$Albert Author is with Faculty of Electrical Engineering, Mathematics and Computer Science,
        University of Twente, 7500 AE Enschede, The Netherlands
        {\tt\small albert.author@papercept.net}}%
\thanks{$^{2}$Bernard D. Researcheris with the Department of Electrical Engineering, Wright State University,
        Dayton, OH 45435, USA
        {\tt\small b.d.researcher@ieee.org}}%
}

\twocolumn[{%
\renewcommand\twocolumn[1][]{#1}%
\maketitle
\begin{center}
    \centering
    \captionsetup{type=figure}
    \includegraphics[width=.24\linewidth]{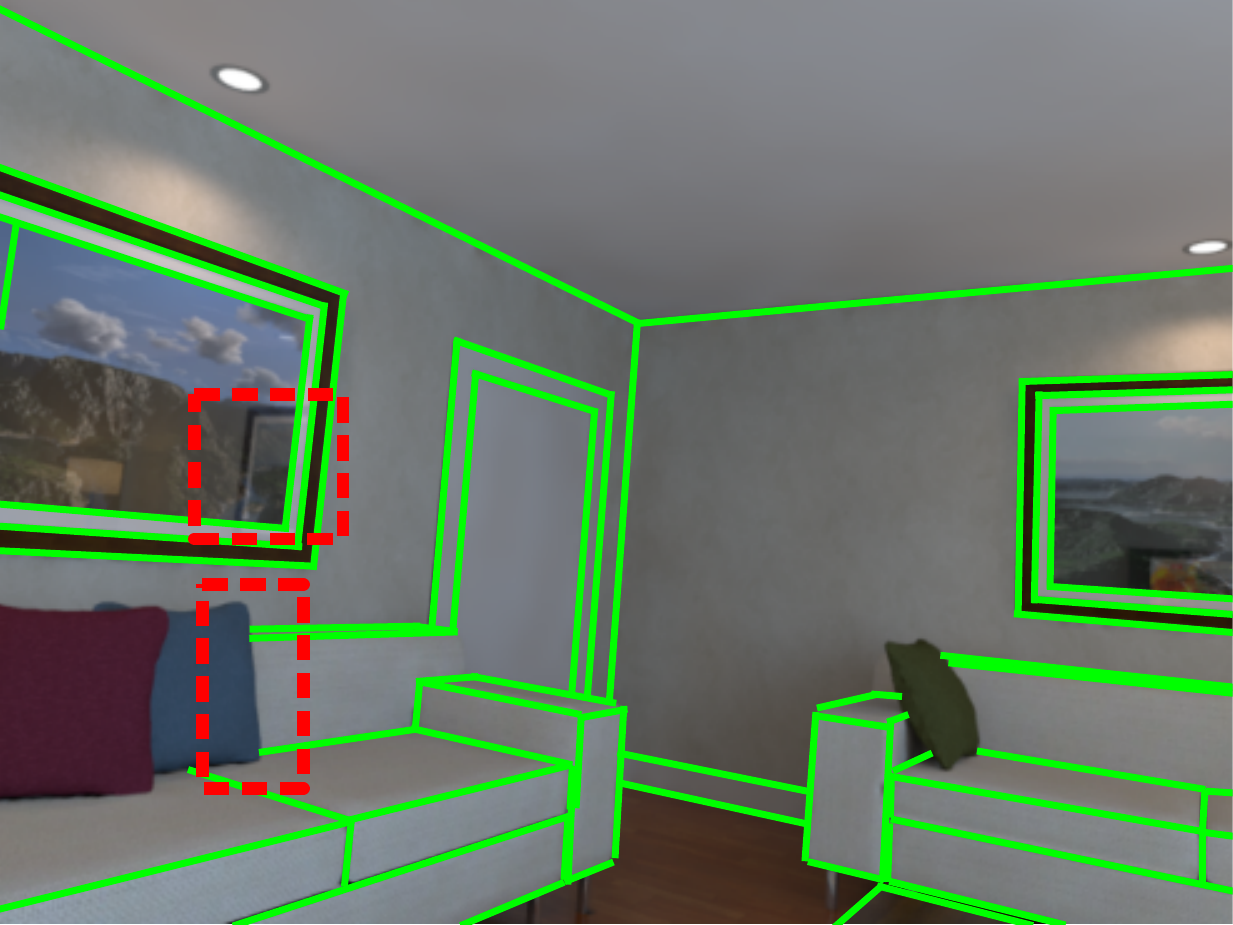}
    \includegraphics[width=.24\linewidth]{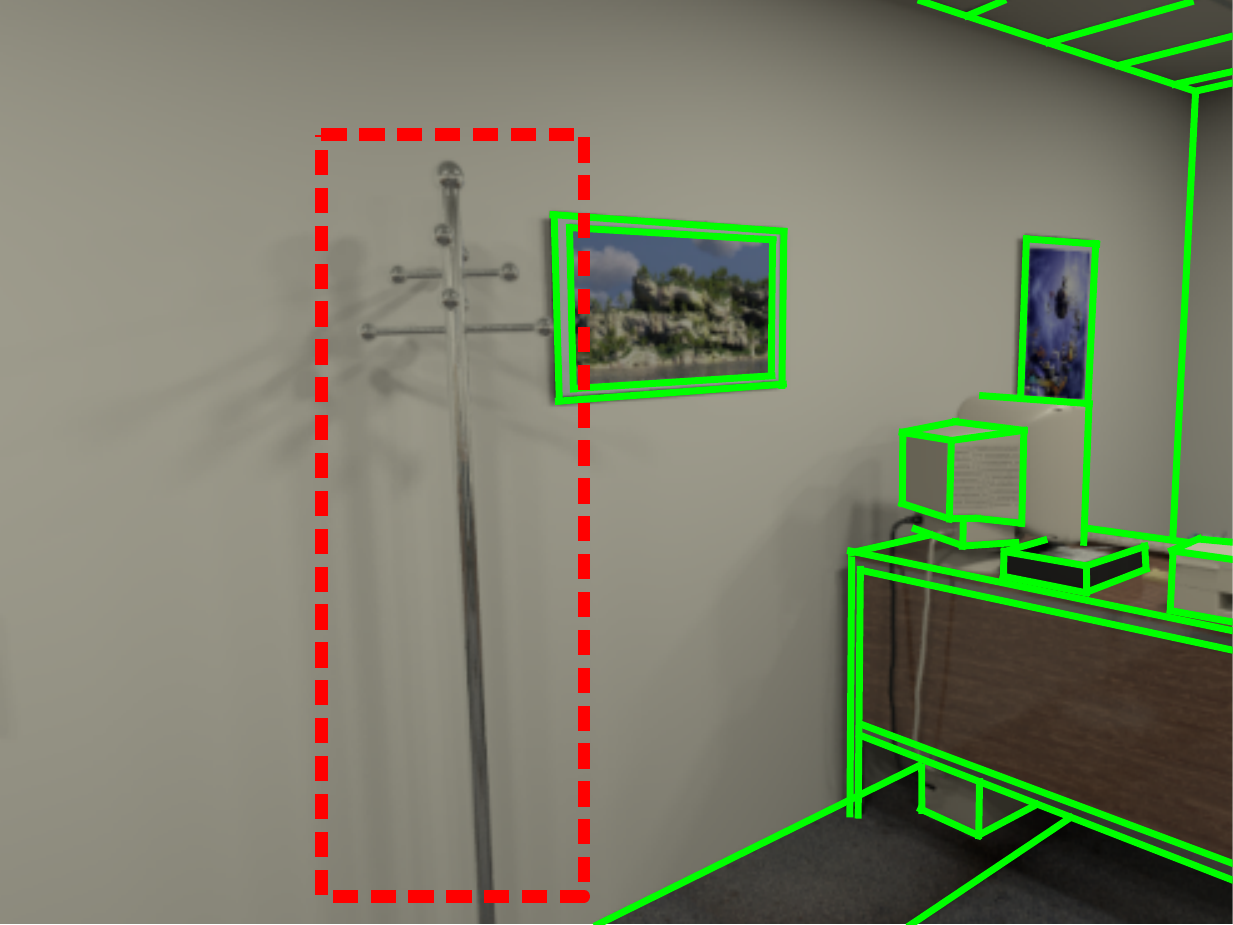}
    \includegraphics[width=.24\linewidth]{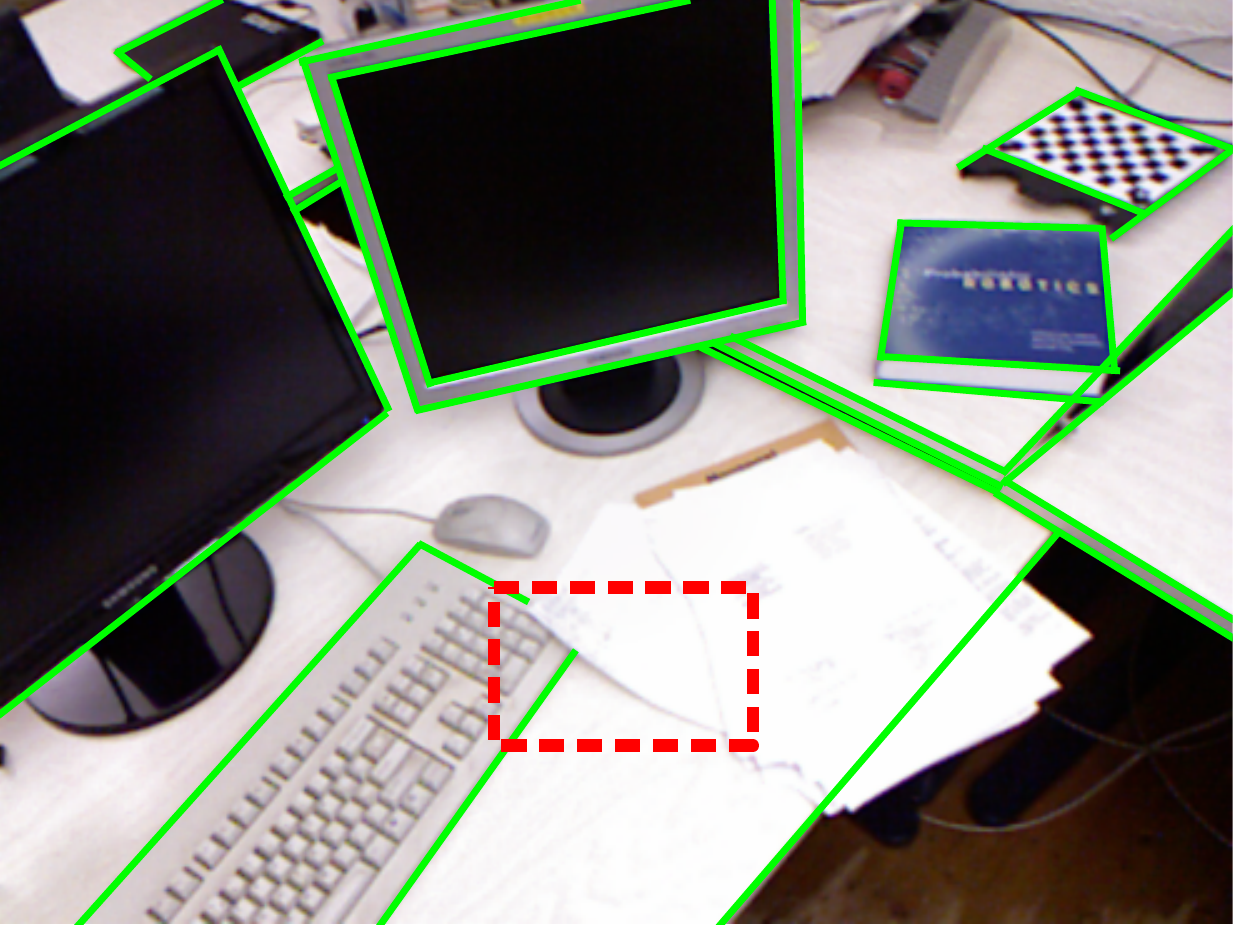}
    \includegraphics[width=.24\linewidth]{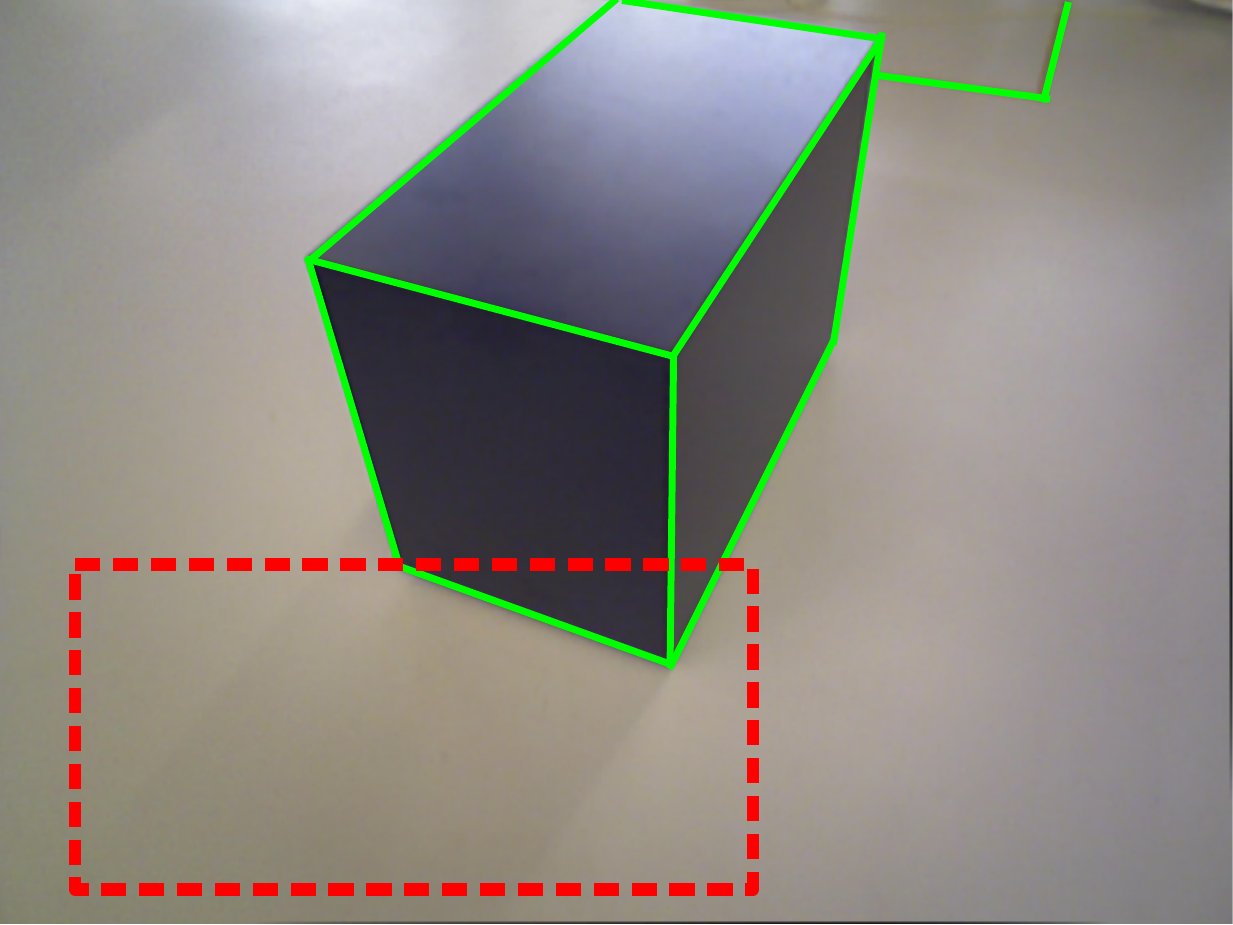}
    \caption{Examples of annotated lines (green) in ICL NUIM and TUM RGBD datasets from our benchmark. Red indicates elements that were not annotated: reflections, elements that form lines from a certain angle, and shadows.}
    \label{fig:teaser}
\end{center}%
}]

\addtocounter{footnote}{1}
\footnotetext{The authors with the Skolkovo Institute of Science and Technology (Skoltech), Center for AI Technology (CAIT).
         {\tt\small {k.ivanov, g.ferrer, anastasiia.kornilova}@skoltech.ru}}
\addtocounter{footnote}{1}
\footnotetext{The authors are with the Software Engineering Department, Saint Petersburg State University.}
\addtocounter{footnote}{1}
\footnotetext{https://prime-slam.github.io/evolin/}

\addtocounter{footnote}{-2}

\begin{abstract}
Lines are interesting geometrical features commonly seen in indoor and urban environments.
There is missing a complete benchmark where one can evaluate lines from a sequential stream of images in all its stages: Line detection, Line Association and Pose error. To do so, we present a complete and exhaustive benchmark for visual lines in a SLAM front-end, both for RGB and RGBD, by providing a plethora of complementary metrics.
We have also labeled data from well-known SLAM datasets in order to have all in one poses and accurately annotated lines. In particular, we have evaluated 17 line detection algorithms, 5 line associations methods and the resultant pose error for aligning a pair of frames with several combinations of detector-association. We have packaged all methods and evaluations metrics and made them publicly available on web-page$^3$.
\end{abstract}
\thispagestyle{empty}
\pagestyle{empty}



\section{Introduction}


Landmark recognition is an essential part of many computer vision algorithms. Thus, the critical task of the Simultaneous Localization and Mapping (SLAM) systems front-end part is the detection of landmarks and their association in a frame stream. In such algorithms, it is popular to use key points as landmarks due to the ease of their recognition. At the same time, point-based SLAM pipelines can produce inaccurate results in poorly textured environments~\cite{gomez2019pl, li2021rgb}. Such environments, however, often contain many lines, which can significantly increase the accuracy and stability of the autonomous system’s motion trajectory estimation; therefore, lines have recently been actively used in SLAM.

Currently, most line-based SLAM frameworks~\cite{pumarola2017pl, zuo2017robust, liu2018stereo, wang2018improved, gomez2019pl, fu2019robust, lee2019elaborate, li2020structure, li2021rgb, lee2021plf, yunus2021manhattanslam, shu2022structure} use the LSD~\cite{von2008lsd} algorithm or its modifications for detection and LBD~\cite{zhang2013efficient} or its variations for association. Their popularity can be explained in terms of their speed. At the same time, many detectors~\cite{akinlar2011edlines, lee2014outdoor, almazan2017mcmlsd, suarez2018fsg, suarez2022elsed, huang2018learning, xue2019learning, zhou2019end, zhang2019ppgnet, lin2020deep, huang2020tp, xue2020holistically, xu2021line, gu2021towards, li2021ulsd, teplyakov2022lsdnet, dai2022fully} and associations~\cite{li2015robust, li2016hierarchical, jia2016novel, vakhitov2019learnable, lange2019dld, lange2020wld, ma2020robust, chen2021hierarchical, yoon2021line, qiao2021superline} have appeared in recent years, including neural networks, which, however, are not actively used in SLAM systems.

The choice of optimal algorithms for use in the SLAM system is problematic, primarily due to the lack of a universal basis for assessing the quality of detection and association, including testing them on famous SLAM sequences. Also, algorithms are often evaluated on data not initially intended for SLAM systems.

This paper provides a universal framework for benchmarking line detection and association algorithms in the SLAM problem. It includes a dataset containing line-annotated images based on known SLAM sequences and metrics for evaluating detection and association algorithms, including line-based relative pose estimation metrics. In addition, we provide a detailed evaluation of popular line detection and association algorithms. Overall, our main contribution is as following.
\begin{itemize}
    \item Labeled dataset with lines on popular SLAM sequences.
    \item Set of docker containers with available detection and association algorithms for lines.
    \item Python library with metrics for detection, association and pose estimation.
    \item Accuracy and performance evaluation of existing approaches to detection and association on labeled sequences.
\end{itemize}

\section{Related work}
In our literature review, we tackle the next topics related to line usage in odometry and SLAM pipelines: detection and association algorithms approaches to use lines in SLAM pipelines, popular metrics, and available datasets with lines.


\textbf{Line Detectors.} The following classification can be distinguished  \cite{rahmdel2015review}: (i) approaches based on the Hough transform, (ii) local approach, (iii) combined approach, and (iv) neural network approaches. 

\textit{The Hough transform}~\cite{illingworth1988survey} consists of the transition from the image space to the parameter space and the subsequent search for extreme values corresponding to lines using the voting procedure. Along with the classical Hough transform~\cite{duda1972use}, there are a large number of optimizations that include probabilistic algorithms~\cite{stephens1991probabilistic, matas2000robust, fernandes2008real}, as well as algorithms that analyze neighborhoods of peak values in the parametric space~\cite{furukawa2003accurate}. 

Algorithms of the \textit{bottom-up approach}, also known as the local approach, exploit the next strategy: first, a primary set of small linear segments are segmented, then they are iteratively linked into larger ones based on the analysis of neighboring elements, optionally having last step to filter our false segments. As the authors of the work~\cite{von2008lsd} showed, early detectors~\cite{burns1986extracting, boldt1989token} performed slowly and recognized many insignificant line segments. In subsequent works, the mechanisms for controlling false alarms have been optimized~\cite{von2008lsd, suarez2018fsg}, and a significant acceleration of the detectors is achieved~\cite{akinlar2011edlines, suarez2022elsed}.

\textit{Combined approach} algorithms use information obtained from the input image and the parametric Hough transform space, which accelerates the voting process~\cite{song2005hough}, selects likely candidates for the local approach~\cite{bandera2006mean}, and filters false positives~\cite{almazan2017mcmlsd}.

Thanks to the development of deep learning in the last decade, multiple \textit{neural network algorithms} have appeared for detecting line segments. Various methods have been proposed for recognizing line segments in images, such as extracting them from attraction field maps~\cite{xue2019learning} and point-pair graph-based detection~\cite{zhang2019ppgnet}. Different architectures have been proposed for line detection, such as transformer-based~\cite{xu2021line}, fully convolutional~\cite{dai2022fully}, and having a trainable module that performs the Hough transform~\cite{lin2020deep}. Various line representations have also been proposed, such as tri-point~\cite{huang2020tp} and Bezier curves~\cite{li2021ulsd}. In addition, real-time detectors have been proposed~\cite{gu2021towards, teplyakov2022lsdnet}. The wireframe parsing problem is also formulated~\cite{huang2018learning, zhou2019end, xue2020holistically}, consisting of finding a set of segments in the image and their junctions.

\textbf{Line Association.}
The following approaches can be distinguished \cite{li2016line}:
(i) line descriptor, (ii) a junction-based approach, (iii) invariants between lines and points, (iv) line segment grouping, and (v) neural networks. 


\textit{A line segment descriptor} is usually represented as a fixed-dimensional vector. An association between segments is established if the distance between the corresponding descriptors is less than a given threshold. In early works~\cite{medioni1985segment, schmid1997automatic}, this approach was used to construct association algorithms, while the descriptor could include information about the gradient, intensity, and color in the neighborhood of the lines. In subsequent works~\cite{bay2005wide, wang2009msld, zhang2013efficient}, optimizations are applied to ensure the resilience of descriptors to changes in lighting and scale.

\textit{Junction-based association} matches segments in pairs based on their intersection points. Thus, in \cite{kim2010novel}, the authors use the coplanarity of pairs of intersecting lines to separate true matches from false ones. In \cite{li2015robust}, a ray-point-ray descriptor is proposed, aggregating information about the lines and their intersection point. The authors of the work~\cite{li2016hierarchical} propose a hierarchical two-stage algorithm that matches line-junction-line descriptors at the first stage and descriptors of individual lines at the second.

\textit{Line association using invariants between coplanar points and line segments}~\cite{fan2012robust, jia2016novel} usually establishes correspondences between points by well-known algorithms. After that, line segments that satisfy the invariants are considered true matches. 

\textit{Algorithms using the grouping approach} are based on creating groups of line segments according to some criterion with their subsequent comparison. Thus, it is proposed to form groups of linear segments lying next to each other based on the similarity of the average gradient values of all pixels of the segments~\cite{wang2009wide}. The authors of work~\cite{chen2021hierarchical} propose dividing all detected line segments into three hierarchical groups and then establishing correspondences sequentially for each group using information obtained from the associations of the previous group.

In recent years, \textit{neural-based} association has also appeared in addition to detectors. In such algorithms, neural network architectures based on convolutional layers are the most popular~\cite{vakhitov2019learnable, lange2019dld, lange2020wld}. Lines can also be represented as sequences of points, allowing techniques from natural language processing~\cite{yoon2021line} and point association~\cite{pautrat2021sold2} to be applied.

\textbf{Line-based SLAM.} In the last years, a set of SLAM algorithms that exploit lines in graph optimizations have been presented by the community~\cite{zhang2015building, zuo2017robust, pumarola2017pl, lee2018monocular, gomez2019pl, fu2019robust, li2020structure, fu2020pl, li2021rgb, lee2021plf, yunus2021manhattanslam}. They can be categorized by approaches used in SLAM front-end and back-end. The common choice for SLAM front-end, namely for detection and association, is LSD detector~\cite{von2008lsd} and LBD association~\cite{zhang2013efficient}. Back-end approaches can be differed by the way of line representation and observation function. The most commonly used representations are Plucker coordinates and line end points. Using this, different variations of observation functions are provided based on reprojection error: end points error, point-to-line error. 


  

\textbf{Datasets.} Two widely used datasets for line detection evaluation~--- YorkUrban~\cite{denis2008efficient} and ShanghaiTech Wireframe~\cite{huang2018learning}. They contain accurate annotations for lines but do not provide sequential frames, making them difficult to apply to the line association problem. In order to test line association, several approaches are used, such as projective transformations of original images~\cite{pautrat2021sold2, qiao2021superline}, automatic annotation~\cite{yoon2021line, lange2019dld}, or referenceless comparison on successive frames~\cite{vakhitov2019learnable}.

\textbf{Metrics.} Line segments can be represented in vector form, i.e., endpoints, and raster form, in the form of a set of points, making it possible to evaluate two types of detection metrics. In the vector case, one can evaluate \textit{the structural average precision} ($sAP$) and \textit{the structural F-score}~\cite{zhou2019end, lin2020deep, xue2020holistically, xu2021line}. Also, in work~\cite{pautrat2021sold2}, repeatability metrics are proposed that reflect the proportion of the same lines detected on different frames. In the raster case, it is possible to calculate the average heatmap precision ($AP^H$) and heatmap F-score ($F^H$)~\cite{martin2004learning, huang2018learning}. Like detection, line association can be viewed as a classification task~\cite{pautrat2021sold2, lange2019dld, lange2020wld} based on which known metrics such as \textit{precision}, \textit{recall}, \textit{F-score}, and others are applied. In addition, it is possible to evaluate the performance of association using metrics for visual localization and homography estimation~\cite{yoon2021line} and, by integrating into the SLAM pipeline, to evaluate the resulting trajectory~\cite{vakhitov2019learnable}.

\section{Data Annotation}
Existing datasets with lines~\cite{denis2008efficient, huang2018learning} do not provide a labeled frame stream with line labels and associations that is required for SLAM evaluation (front-end and back-end). This section describes the choice of datasets for annotating lines within our benchmark and the annotating process.

\subsection{Dataset Choice}
Popular datasets for evaluating line detectors and association do not contain consecutive frames, which makes it difficult to assess their applicability in SLAM pipelines. 
To solve this problem, we choose the ICL-NUIM~\cite{handa2014benchmark} and TUM RGB\mbox{-}D~\cite{sturm2012benchmark} datasets for line segment annotation, which contain consecutive frames and are actively used to evaluate the performance of SLAM systems. The choice of these datasets is also due to the fact that they are already annotated with planes~\cite{kornilova2022evops}, which allows the creation of a joint framework for testing lines, points, and planes in the front-end of SLAM. We take trajectories \texttt{lr kt2} and \texttt{of kt2} from ICL NUIM that scan the whole scene, and trajectories \texttt{fr3/cabinet} and \texttt{fr1/desk} from TUM RGB-D for annotating.

\subsection{Annotation process}

The above-specified datasets are annotated manually using the CVAT tool~\cite{boris_sekachev_2020_4009388}, which allows tracking lines on successive frames, which is important for obtaining ground truth associations. The annotation process consisted of the following steps: defining lines on the first frame, interpolating lines on subsequent frames using the CVAT functionality, and correcting lines on intermediate frames. If the line is absent on the scene, it is not displayed in the track on the corresponding frames. Only breaking segments have been annotated, such as ceilings, floors, walls, doors, and furniture linear elements. Examples of different scene annotations can be seen in Fig.~\ref{fig:teaser}. Linear locally and texture elements are not annotated. The statistics of annotated lines among labeled datasets are presented in Tab.~\ref{table:data_stats}.


\begin{table}[!htb]
\centering
\caption{Statistics on annotated line segments on selected datasets}
\begin{tabular}{ | p{.2\linewidth} || p{.15\linewidth} | p{.15\linewidth} | p{.2\linewidth} | }
 \hline
 Dataset & \# lines & \# frames & Lines per Frame\\
 \hline
 lr kt2 & 189 & 881 & 47 \\
 of kt2 & 346 & 881 & 78 \\
 fr3/cabinet & 46 & 1147 & 13 \\
 fr1/desk & 184 & 613 & 51 \\
 \hline
\end{tabular}
\label{table:data_stats}
\end{table}

\section{Metrics}
This section describes the detection and association metrics used in our benchmark and the relative pose estimation metrics based on line correspondences. We combine them into a single Python-library.

\subsection{Detection}
 One can distinguish between \textit{vectorized} and \textit{heatmap-based} metrics depending on the line representation. In the first, line segments are defined by their endpoints, and in the second, by rasterized points.
 
\textbf{Vectorized metrics} include adapted versions of classification~\cite{zhou2019end} and repeatability~\cite{pautrat2021sold2} metrics. Both groups of metrics require a distance function $d(l_1, l_2)$,  where $l_i = (p_i^1, p_i^2)$ is an $i$-th line segment defined by its endpoints $p_i^1, p_i^2 \in \mathbb{R}^2$. As in work~\cite{pautrat2021sold2}, we use the following distances. 

Structural distance between lines $i$ and $j$: 
\begin{equation}
\begin{aligned}
    d_s(l_i, l_j) = \min\{ & ||{p}^{1}_j-p_i^1||_2 + || {p}^{2}_j-p_i^2||_2, \\
        & || {p}^{1}_j-p_i^2||_2+ || {p}^{2}_j-p_i^1||_2\}. \label{eq:struct_dist}
\end{aligned}
\end{equation}

Orthogonal distance between lines $i$ and $j$:
    \begin{equation}
        d_{\perp}(l_i, l_j) = \dfrac{d_a(l_i, l_j) + d_a(l_j, l_i)}{2}, \label{eq:orth_dist}
    \end{equation}
    where 
    \begin{equation}
        d_a(l_i, l_j) = \lVert p_j^1 - \pi_{l_i}(p_j^1) \rVert_2 + \lVert p_j^2 - \pi_{l_i}(p_j^2) \rVert_2,
    \end{equation}
    and $\ \pi_l(\cdot)$ is the orthogonal projection onto the line $l$.

\paragraph{Classification metrics}
A detected line segment $l_j$ is considered True Positive (TP) if and only if the following conditions are met:
\begin{itemize}
    \item $\min_{l \in GT}d(l_j, l) \le d_{max}$, where $d_{max}$ is the maximum allowable distance between the predicted and reference segments expressed in pixels, and $GT$ is the set of ground truth line segments;
    \item if there exists $l_i$ such that $\arg \min_{l \in GT}d(l_i, l) = \arg \min_{l \in GT}d(l_j, l) = l'$, then $ l_j$ must have a higher detector confidence level than $l_i$ in case the detector provides confidence values for each segment. If the detector does not provide such values, then the inequality $d(l_j, l') < d(l_i, l')$ must hold.
\end{itemize}
After establishing for each predicted line whether it is TP or False Positive (FP), we can calculate classical classification metrics such as \textit{precision}, \textit{recall}, \textit{F-score}, and \textit{average precision}.

\paragraph{Repeatability metrics}
Since the annotated datasets contain information about the depth and intrinsic camera parameters, we explicitly reproject lines from one frame to another rather than using synthetic homographies, as in work~\cite{pautrat2021sold2}. Let $L_1$ and $L_2$ be the sets of lines in the first and second images, respectively, and $P_{i \rightarrow j}(\cdot)$ be the transformation mapping the lines from the $i$-th image to the $j$-th. Let $L_1' = P_{1 \rightarrow 2}(L_1)$ and $L_2' = P_{2 \rightarrow 1}(L_2)$. We also need to define the following indicator that the distance from the line $l$ to the set of lines $L$ is less than the threshold $d_{max}$:
\begin{equation}
        I_{L}(l) = \begin{cases}
                        1, & \text{if } \min_{l_j\in L}d(l, l_j) \leq d_{max}\\
                        0, & \text{otherwise}
                     \end{cases},
\end{equation}
where $d(\cdot, \cdot)$ is either an orthogonal or structural distance, as described earlier.
Now we can define the \textit{repeatability} metric showing the proportion of lines found on both the first and second frames and the \textit{localization error}~--- the average distance between such lines.
\begin{itemize}
    \item Repeatability:
    \begin{equation}
        \text{Rep-}{d_{max}} = \frac{\sum_{l_i \in L_1'}I_{L_2}(l_i) + \sum_{l_j \in L_2'}I_{L_1}(l_j)}{ \lvert L_1 \rvert + \lvert L_2 \rvert }.
    \end{equation}
    \item Localization error:
    Let $Dist(L_{j}, L_{i})$ be the set of closest distances from each line in $L_j$ to some line in $L_i$ within the threshold $d_{max}$, which is defined as follows:
    \begin{equation}
        Dist(L_{j}, L_{i}) = \left\{ \min_{l_i\in L_i}d(l_i, l) \Bigm\vert I_{L_i}(l) = 1, l \in L_j \right\}
    \end{equation}
    Then the \textit{localization error} is defined as follows:
    \begin{equation}
        \text{LE-$d_{max}$} = \dfrac{\sum_{d \in Dist(L_{2}', L_{2}) \cup Dist(L_{1}', L_{1})}d}{\lvert Dist(L_{2}', L_{2}) \rvert + \lvert Dist(L_{1}', L_{1}) \rvert}
    \end{equation}
\end{itemize}
Here, $\lvert \cdot \rvert$ is the set cardinality.

\textbf{Heatmap-based metrics.} In the case of a raster representation of lines, we consider detection to classify each pixel from the point of view of belonging to any line, as in the works~\cite{martin2004learning, huang2018learning}. In this context, a heatmap is a boolean matrix that has the same size as the original image and is obtained by rasterizing the predicted or reference line segments such that each of its elements shows the belonging of the corresponding pixel to any segment. To calculate the metric, it is necessary to build heatmaps of the predicted and reference line segments, then determine which pixels on the predicted heatmap are TP and FP. To do this, it is necessary to solve the minimum weight assignment problem~\cite{martin2004learning}, in which the nodes are pixels, and the weights are the distances between them. Only those elements from the predicted heatmap marked as true and the distance from which some reference pixel is less than the specified threshold $d_{max}$ are included in the consideration. The remaining pixels labeled as true pixels are considered FP. After solving this problem, we have a matrix of indicators, whether the corresponding pixel is TP or not, based on which we can calculate well-known classification metrics.

\subsection{Association}
The problem of finding associations between lines on successive frames can also be considered a classification problem. Given a set of ground truth associations $GT$, we can assign each predicted association to one of the following classes: \textbf{TP} if it is established and belongs to the $GT$ set; \textbf{FP} if it is established and does not belong to the $GT$ set; \textbf{TN} if it is not established and does not belong to the $GT$ set; \textbf{FN} if it is not established and belongs to the $GT$ set.

Popular classification metrics are calculated based on the characteristics obtained, as in the case of detection.

\subsection{Pose estimation error}
\label{sec:pose_est}
We also provide a mechanism to evaluate detectors and association quality for pose estimation problem, in similar manner to keypoint detection and association benchmarks~\cite{yi2018learning, sarlin2020superglue}. To do this, we formulate optimization problem in the way proposed in work~\cite{pumarola2017pl}. Let $P, Q \in \mathbb{R}^3$ be endpoints of line in 3D map, $p, q \in \mathbb{R}^2$~--- its detections on image plane, $\bar p, \bar q \in \mathbb{P}^2$~--- corresponding homogeneous coordinates. Then we define line observation $l \in \mathbb{P}^2$ by its coefficients:

\begin{equation}
    \begin{aligned}
        l = \frac{\bar p \times \bar q}{|\bar p \times \bar q|}
    \end{aligned}
\end{equation}

Then line error could be formulated as:

\begin{equation}
    \begin{aligned}
        e(T) = ||l^\top \pi(TP)|| ^ 2 + ||l^\top \pi(TQ)|| ^ 2,
    \end{aligned}
\end{equation}
where $\pi(\cdot)$ denotes projection of the 3D point to the image plane in homogenous coordinates, $T \in SE(3)$ describes transformation from the map to the observed frame.

It is worth to note that since our annotated dataset contains depth information, it is easy to calculate 3D points corresponding to line edges. Finally, we estimate relative pose $T$ by extending g2o framework~\cite{kummerle2011g} with an edge implementation that describes this observation function. Also, in optimizations robust Huber kernel was used to filter out outliers as done in majority of existing line-based SLAM. When relative pose is estimated, statistics on translation and rotation error can be calculated.

\section{Experiments}
This section provides an evaluation of existing line detection and association algorithms on our line-annotated data.

\subsection{Experimental Settings}

\begin{table*}[!htb]
\centering
\caption{Comparison of detectors with line uncertainty measure: average precision for structural ($sAP$) and orthogonal ($oAP$) distances with a 5 and 10 pixels threshold and FPS.}
\label{table:vectorized_scored}
\huge
\resizebox{1\linewidth}{!}{
\begin{tabular}{|l||llll|l|llll|l|llll|l|llll|l|} 
\hline
\multirow{2}{*}{Algorithm} & \multicolumn{5}{c|}{lr kt2}                                                  & \multicolumn{5}{c|}{of kt2}                                                  & \multicolumn{5}{c|}{fr1/desk}                                               & \multicolumn{5}{c|}{fr3/cabinet}                                              \\ 
\cline{2-21}
                           & $sAP^5$       & $sAP^{10}$    & $oAP^5$       & $oAP^{10}$    & FPS          & $sAP^5$       & $sAP^{10}$    & $oAP^5$       & $oAP^{10}$    & FPS          & $sAP^5$       & $sAP^{10}$    & $oAP^5$       & $oAP^{10}$    & FPS         & $sAP^5$       & $sAP^{10}$    & $oAP^5$       & $oAP^{10}$    & FPS           \\ 
\hline
AFM                        & 32.2          & 41.4          & 54.6          & 56.8          & 12           & 16.0          & 23.3          & 42.9          & 46.2          & 12           & 10.4          & 19.3          & 28.0          & 34.3          & 11          & 39.0          & 49.3          & 66.0          & 70.7          & 10            \\
ELSED                      & 23.2          & 29.3          & 47.4          & 48.6          & \uline{107}  & 14.7          & 22.9          & 40.7          & 42.0          & 98           & 16.0          & 24.4          & 38.1          & 41.1          & \uline{69}  & 56.0          & 63.8          & 76.9          & 77.8          & \textbf{162}  \\
F-Clip                     & 68.6          & 74.0          & 79.5          & 80.0          & 30           & \textbf{67.5} & \textbf{72.4} & \textbf{75.5} & \textbf{77.2} & 29           & \uline{36.3}  & \uline{42.7}  & \uline{46.7}  & \uline{50.5}  & 29          & 87.1          & 87.5          & 90.5          & \uline{90.9}  & 29            \\
HAWP                       & \textbf{73.2} & \textbf{77.8} & \textbf{82.0} & \textbf{82.6} & 25           & \uline{66.3}  & \uline{70.9}  & \uline{73.0}  & \uline{75.0}  & 23           & 36.1          & 42.1          & 45.2          & 49.1          & 23          & \uline{87.7}  & 88.8          & 89.5          & 90.2          & 26            \\
HT-LCNN                    & 68.2          & 73.3          & 76.8          & 77.2          & 7            & 61.6          & 65.9          & 68.4          & 70.3          & 6            & 33.1          & 39.2          & 41.6          & 45.9          & 5           & 82.8          & 83.2          & 84.2          & 84.6          & 8             \\
L-CNN                      & 67.0          & 72.2          & 76.0          & 76.5          & 17           & 61.2          & 66.0          & 67.7          & 69.8          & 11           & 31.7          & 37.5          & 40.0          & 43.7          & 8           & 83.7          & 84.2          & 85.3          & 86.0          & 25            \\
LETR                       & 64.8          & 72.2          & 78.9          & 79.6          & 8            & 60.5          & 69.5          & 72.3          & 74.4          & 8            & \textbf{38.3} & \textbf{49.1} & \textbf{54.0} & \textbf{58.9} & 8           & 87.0          & 88.8          & \uline{91.1}  & \textbf{91.7} & 8             \\
LSD                        & 17.3          & 23.1          & 44.8          & 46.3          & \textbf{111} & 12.8          & 19.9          & 41.6          & 43.7          & \textbf{112} & 8.3           & 16.7          & 31.5          & 35.6          & 61          & 30.9          & 39.8          & 60.0          & 61.6          & \uline{139}   \\
LSDNet                     & 33.3          & 42.5          & 54.4          & 55.7          & 13           & 13.0          & 19.5          & 36.2          & 39.7          & 13           & 12.7          & 22.7          & 31.8          & 37.1          & 13          & 54.0          & 60.3          & 74.4          & 75.9          & 13            \\
LSWMS                      & 1.5           & 3.9           & 10.7          & 14.6          & 20           & 2.9           & 8.2           & 20.5          & 27.7          & 25           & 1.5           & 5.2           & 11.1          & 17.3          & 20          & 1.8           & 3.3           & 10.3          & 14.2          & 19            \\
M-LSD                      & 60.3          & 67.8          & 74.4          & 75.3          & 98           & 51.9          & 58.3          & 65.5          & 68.5          & \uline{103}  & 33.9          & 40.8          & 44.7          & 48.8          & \textbf{99} & 84.8          & 86.1          & 89.4          & 89.9          & 111           \\
TP-LSD                     & 70.1          & \uline{77.5}  & \uline{81.6}  & \uline{82.2}  & 17           & 58.1          & 64.5          & 68.2          & 70.4          & 17           & 35.3          & 41.7          & 44.8          & 48.5          & 17          & \textbf{88.1} & \textbf{89.0} & \textbf{91.2} & \textbf{91.7} & 17            \\
ULSD                       & \uline{71.4}  & 76.4          & 79.6          & 80.1          & 27           & 62.5          & 67.3          & 70.9          & 72.3          & 26           & 34.1          & 40.4          & 43.4          & 47.3          & 25          & 87.6          & \uline{88.9}  & 89.4          & 90.3          & 28            \\
\hline
\end{tabular}
}

\end{table*}
\begin{table*}[!htb]
\centering
\caption{Comparison of detectors without line uncertainty measure: precision ($P$), recall ($R$), and F-score ($F$) for structural and orthogonal distances with a 5 pixels threshold and FPS.}
\label{table:vectorized_unscored}
\huge
\resizebox{1\linewidth}{!}{
\begin{tabular}{|l||lll|lll|l|lll|lll|l|lll|lll|l|lll|lll|l|} 
\hline
\multirow{3}{*}{Algorithm} & \multicolumn{7}{c|}{lr kt2}                                                                                          & \multicolumn{7}{c|}{of kt2}                                                                                          & \multicolumn{7}{c|}{fr1/desk}                                                                                        & \multicolumn{7}{c|}{fr3/cabinet}                                                                                      \\ 
\cline{2-29}
                           & \multicolumn{3}{c|}{$d_s$}                    & \multicolumn{3}{c|}{$d_{\perp}$}              & \multirow{2}{*}{FPS} & \multicolumn{3}{c|}{$d_s$}                    & \multicolumn{3}{c|}{$d_{\perp}$}              & \multirow{2}{*}{FPS} & \multicolumn{3}{c|}{$d_s$}                    & \multicolumn{3}{c|}{$d_{\perp}$}              & \multirow{2}{*}{FPS} & \multicolumn{3}{c|}{$d_s$}                    & \multicolumn{3}{c|}{$d_{\perp}$}              & \multirow{2}{*}{FPS}  \\ 
\cline{2-7}\cline{9-14}\cline{16-21}\cline{23-28}
                           & $P^5$         & $R^5$         & $F^5$         & $P^5$         & $R^5$         & $F^5$         &                      & $P^5$         & $R^5$         & $F^5$         & $P^5$         & $R^5$         & $F^5$         &                      & $P^5$         & $R^5$         & $F^5$         & $P^5$         & $R^5$         & $F^5$         &                      & $P^5$         & $R^5$         & $F^5$         & $P^5$         & $R^5$         & $F^5$         &                       \\ 
\hline
EDLines                    & \uline{12.1}  & \textbf{56.2} & \uline{19.9}  & 19.2          & \textbf{89.0} & 31.6          & \textbf{387}         & \uline{14.7}  & \textbf{50.8} & \uline{22.8}  & 24.0          & \textbf{82.5} & \uline{37.1}  & \textbf{325}         & 4.2           & \textbf{50.9} & \uline{7.8}   & 7.7           & \textbf{93.2} & 14.3          & \textbf{213}         & 12.9          & \textbf{68.0} & 21.7          & 18.0          & \textbf{94.5} & 30.2          & \textbf{679}          \\
FLD                        & 10.5          & \uline{45.6}  & 17.0          & 18.9          & \uline{82.3}  & 30.7          & \uline{346}          & 13.1          & \uline{45.7}  & 20.4          & 22.7          & \uline{79.1}  & 35.3          & \uline{284}          & 4.0           & \uline{42.9}  & 7.3           & 8.3           & \uline{89.9}  & 15.2          & \uline{180}          & 13.4          & 54.6          & 21.5          & 21.9          & \uline{89.2}  & 35.2          & \uline{668}           \\
FSG                        & \textbf{28.8} & 43.4          & \textbf{34.6} & \textbf{42.1} & 63.4          & \textbf{50.6} & 39                   & \textbf{26.7} & 28.0          & \textbf{27.3} & \textbf{51.5} & 54.0          & \textbf{52.7} & 38                   & \textbf{11.0} & 41.6          & \textbf{17.4} & \uline{19.3}  & 73.0          & \textbf{30.5} & 21                   & \textbf{38.7} & \uline{63.4}  & \textbf{48.0} & \textbf{50.1} & 82.2          & \textbf{62.3} & 51                    \\
SOLD                       & 8.8           & 24.7          & 13.0          & \uline{23.7}  & 66.5          & \uline{35.0}  & 3                    & 11.5          & 27.4          & 16.2          & \uline{24.9}  & 59.3          & 35.1          & 3                    & \uline{6.5}   & 9.3           & 7.7           & \textbf{22.4} & 32.3          & \uline{26.5}  & 4                    & \uline{26.5}  & 42.8          & \uline{32.8}  & \uline{40.0}  & 64.5          & \uline{49.4}  & 7                     \\
\hline
\end{tabular}
}
\end{table*}

\textbf{Hardware.} For the evaluations, a machine with the following characteristics was used:  11th Gen Intel(R) Core(TM) i5-11300H @ 3.10GHz, 24 GB RAM, GeForce RTX 3060 Mobile 6 GB. 

\textbf{Line Detectors.} We have compared and evaluated a total of 17 algorithms: 
AFM~\cite{xue2019learning}, ELSED~\cite{suarez2022elsed}, F-Clip~\cite{dai2022fully}, HAWP~\cite{xue2020holistically}, HT-LCNN~\cite{lin2020deep}, L-CNN~\cite{zhou2019end}, LETR~\cite{xu2021line}, LSDNet~\cite{teplyakov2022lsdnet}, LSWMS~\cite{nieto2011line}, M-LSD~\cite{gu2021towards}, TP-LSD~\cite{huang2020tp}, ULSD~\cite{li2021ulsd}, FSG~\cite{suarez2018fsg}, and SOLD2~\cite{pautrat2021sold2}. Implementations of the following algorithms were taken from OpenCV: LSD~\cite{von2008lsd}, EDLines~\cite{akinlar2011edlines}, and FLD~\cite{lee2014outdoor}.
All methods are packaged into docker containers.

\textbf{Line Association.} For comparison, we select five association algorithms with an open implementation into docker containers: LBD~\cite{zhang2013efficient}, DLD~\cite{lange2019dld}, WLD~\cite{lange2020wld}, LineTR~\cite{yoon2021line}, and SOLD2~\cite{yoon2021line}. We use brute force matching of descriptors for LBD, DLD, and WLD and matching algorithms proposed by the authors for LineTR and SOLD2.

\textbf{Datasets.} For comparison, we use all frames of line-annotated trajectories in \texttt{lr kt2}, \texttt{of kt2} from ICL NUIM, and \texttt{fr3/cabinet}, \texttt{fr1/desk} from TUM RGB\mbox{-}D.

\subsection{Line Detection}

Among the metrics built on the vector representation of the line, for algorithms that provide a measure of uncertainty for each line, we choose the average precision ($AP$) for distances $d_s$ and $d_{\perp}$, from (\ref{eq:struct_dist}) and (\ref{eq:orth_dist}), with a threshold of 5 and 10 pixels and also built PR curves and compared the rest of the algorithms using precision, recall, and F-score. We scale the lines to the $128\times128$ resolution to reduce the impact of image resolution. Among the heatmap-based metrics, we have taken average precision and PR curves for algorithms that measure line uncertainty and precision, recall, and F-score for the rest of the algorithms. We also compute the repeatability metrics for the previously specified distances with a threshold of 5 pixels on the datasets. These metrics are not calculated on the \texttt{TUM fr1/desk} due to the poor quality of the depth maps. In this case, we filter the line segments output by the detector with the threshold of the uncertainty measure if it is available.
 
\begin{figure*}[!htb]
     \centering
     \includegraphics[width=.85\linewidth]{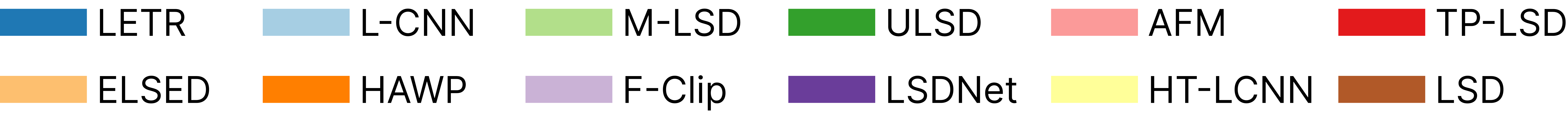}
     \subfloat[lr kt2]{\includegraphics[width=.245\linewidth]{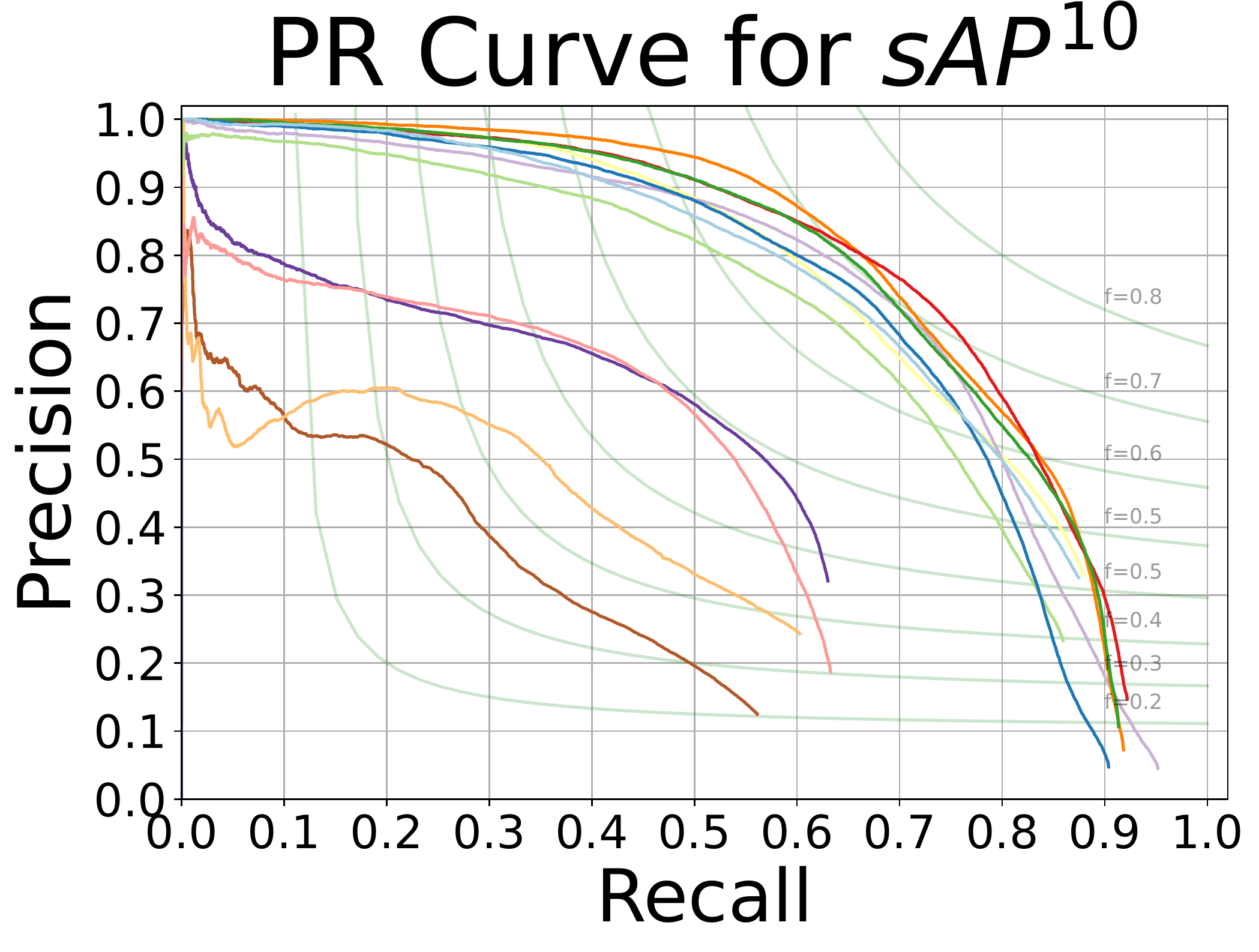}}
     \subfloat[of kt2]{\includegraphics[width=.245\linewidth]{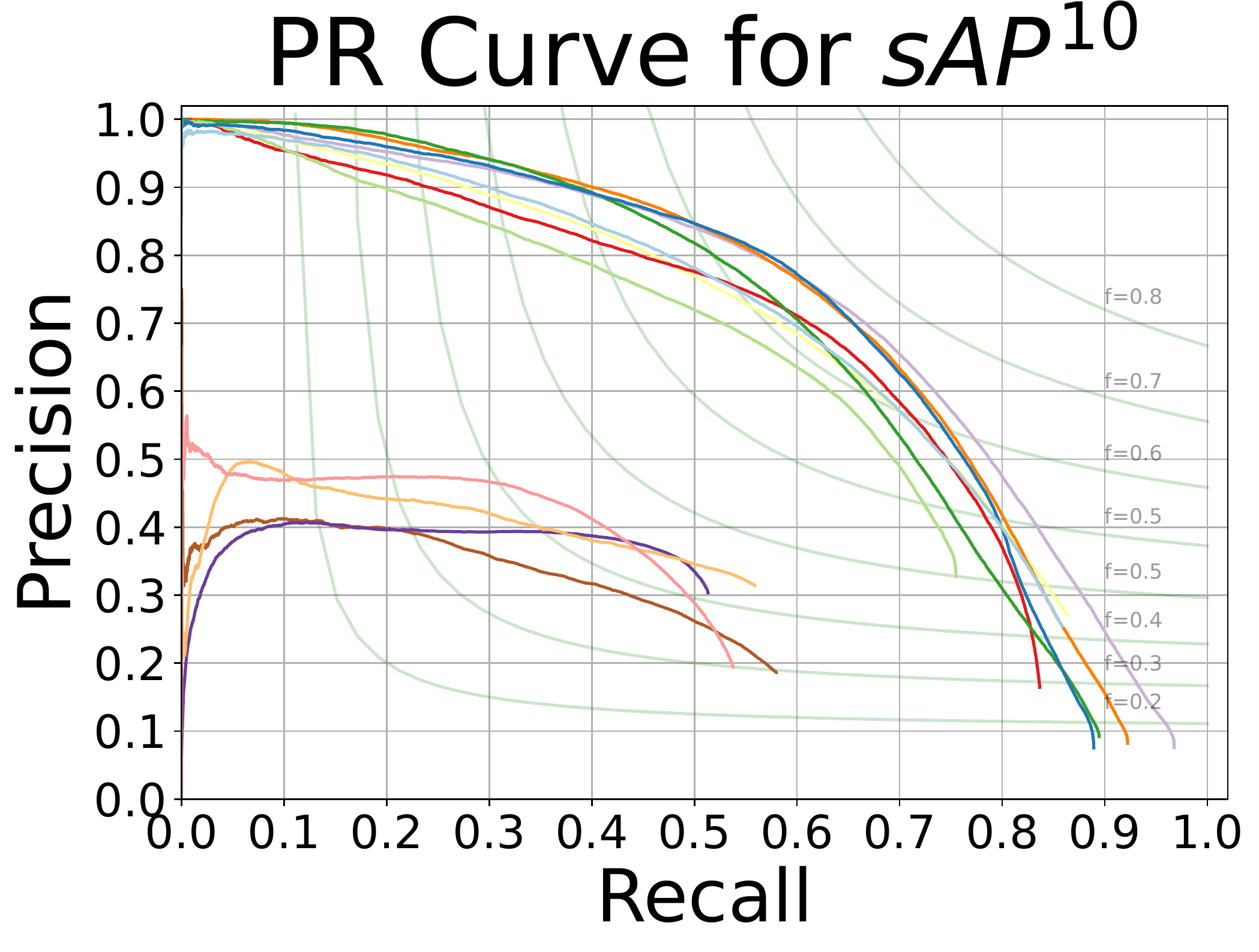}}
     \subfloat[fr1/desk]{\includegraphics[width=.245\linewidth]{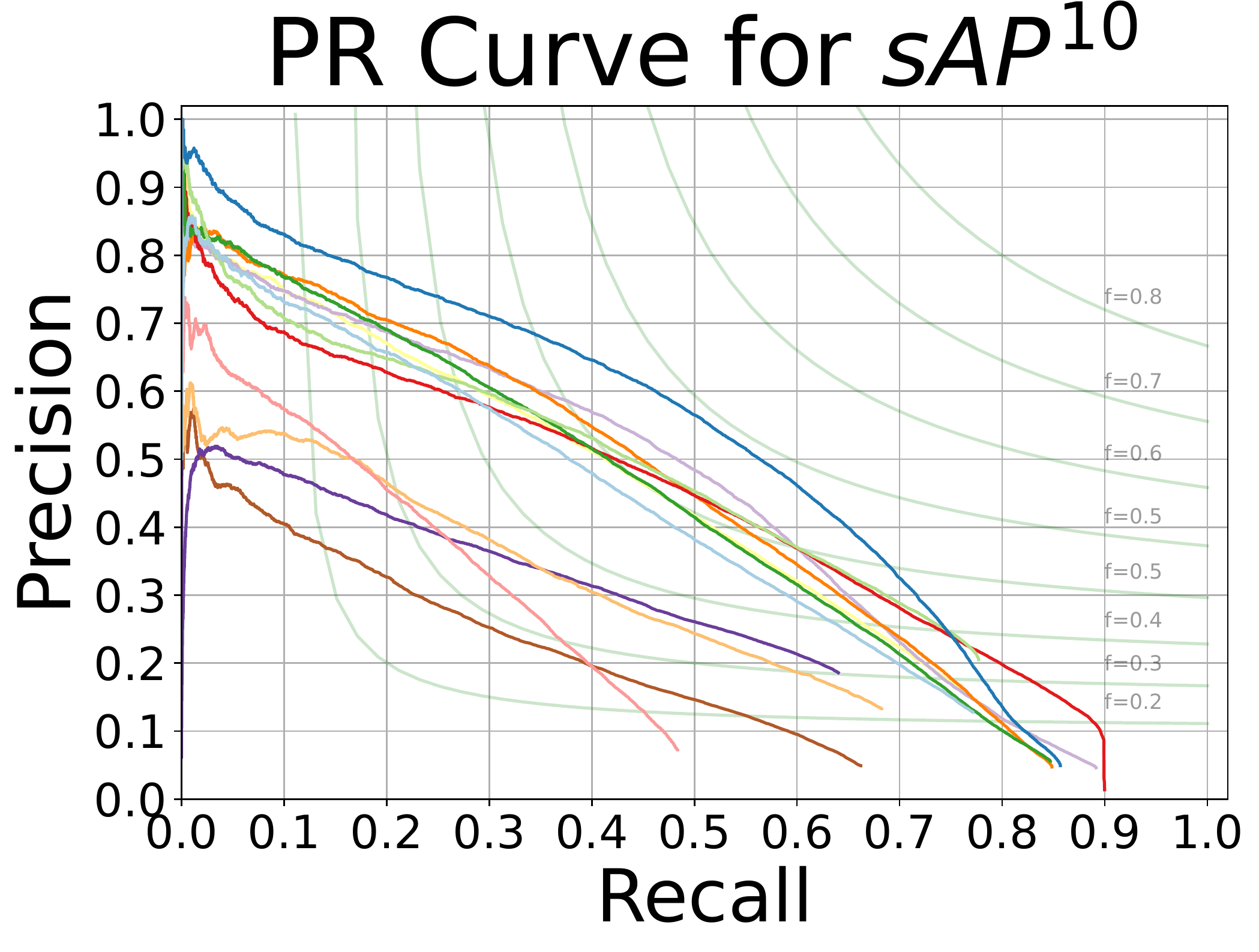}}
     \subfloat[fr3/cabinet]{\includegraphics[width=.245\linewidth]{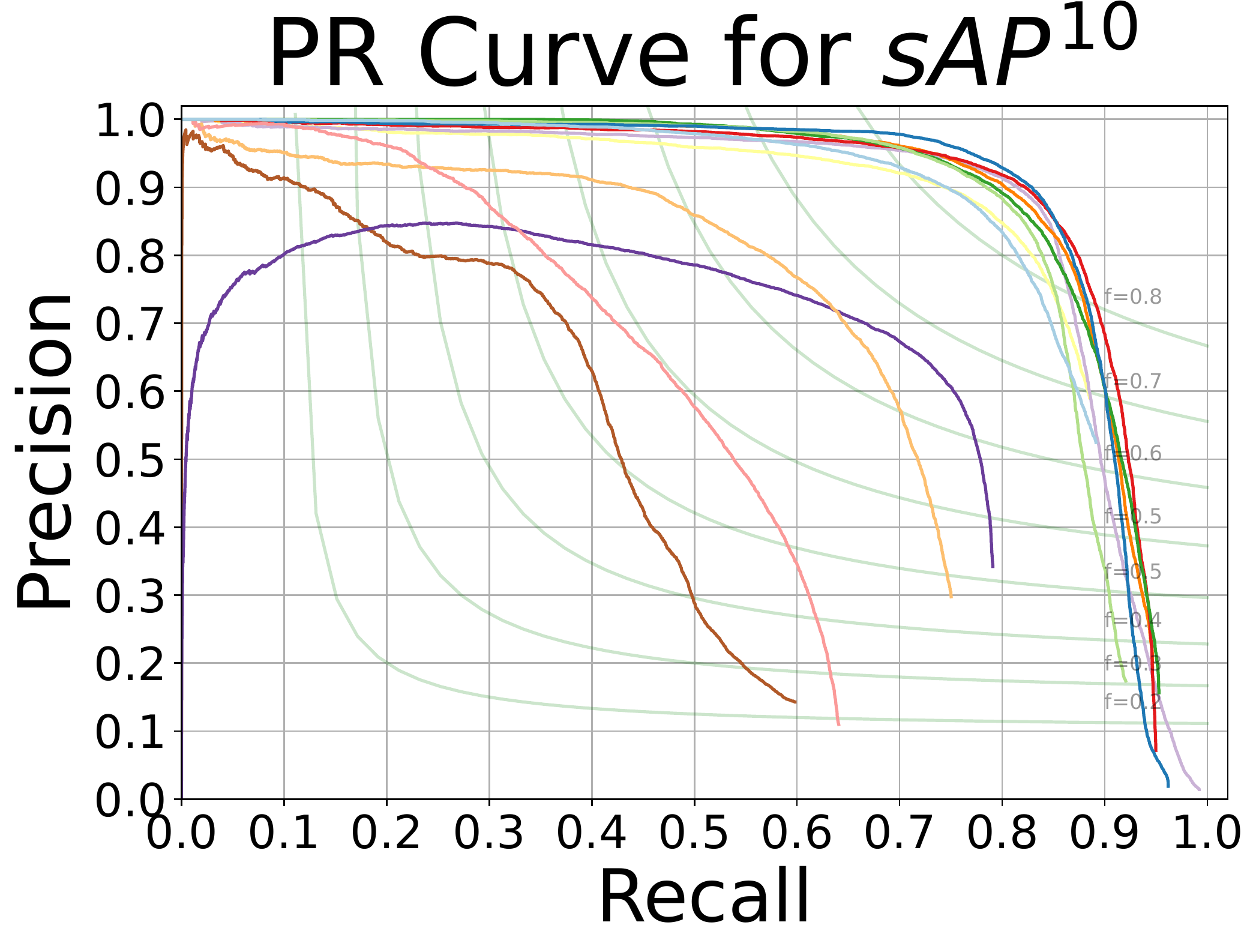}}\qquad
      \subfloat[lr kt2]{\includegraphics[width=.245\linewidth]{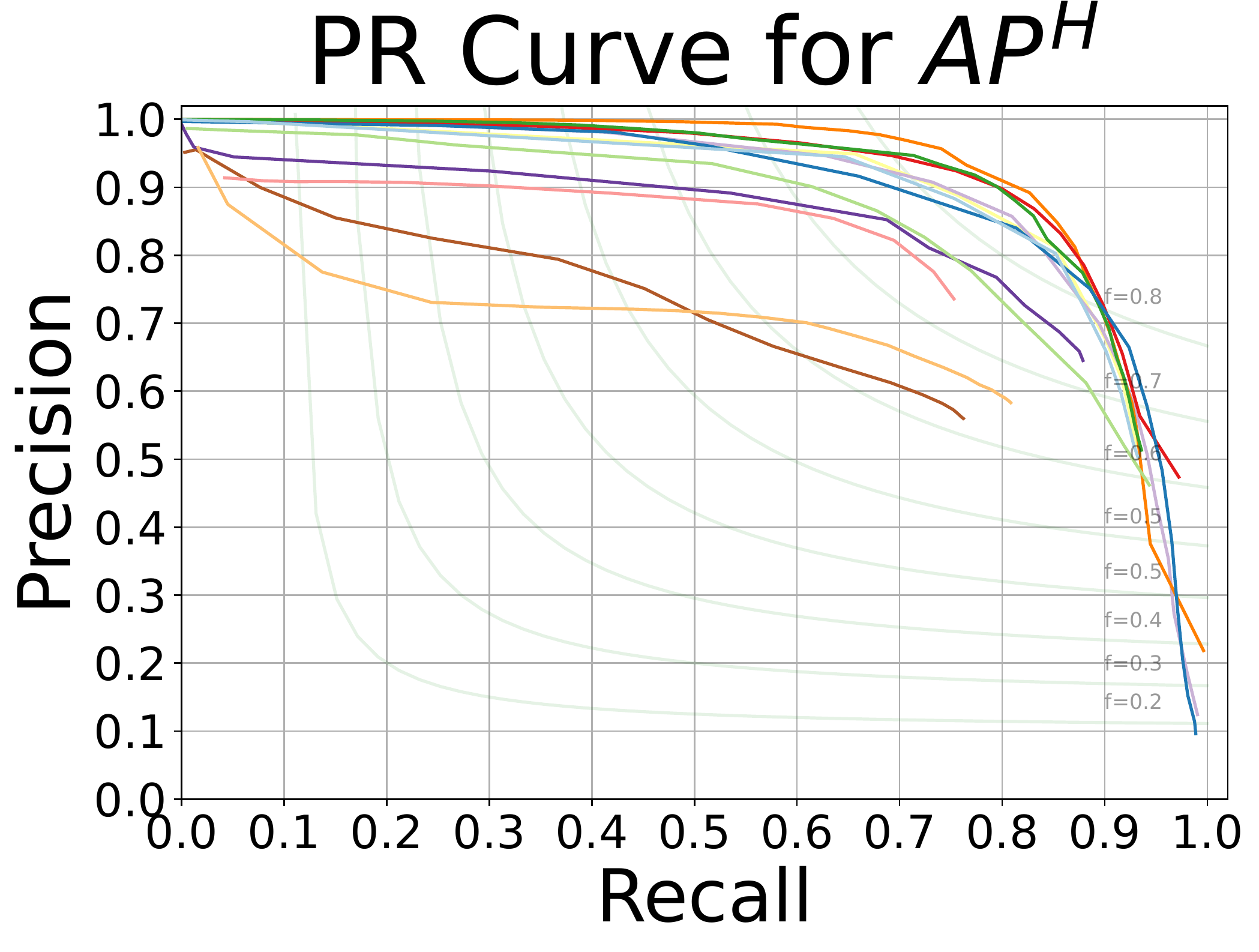}}
     \subfloat[of kt2]{\includegraphics[width=.245\linewidth]{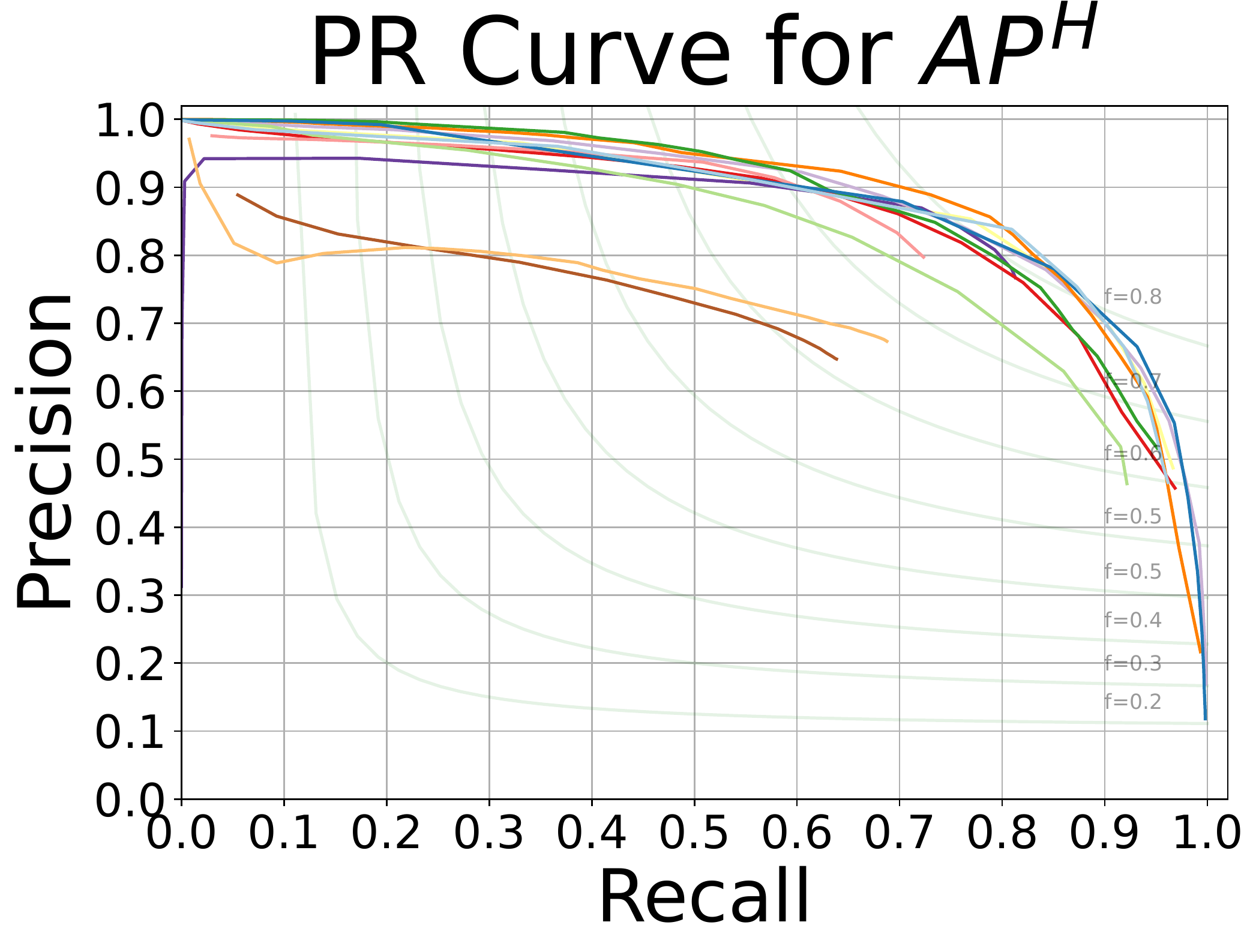}}
     \subfloat[fr1/desk]{\includegraphics[width=.245\linewidth]{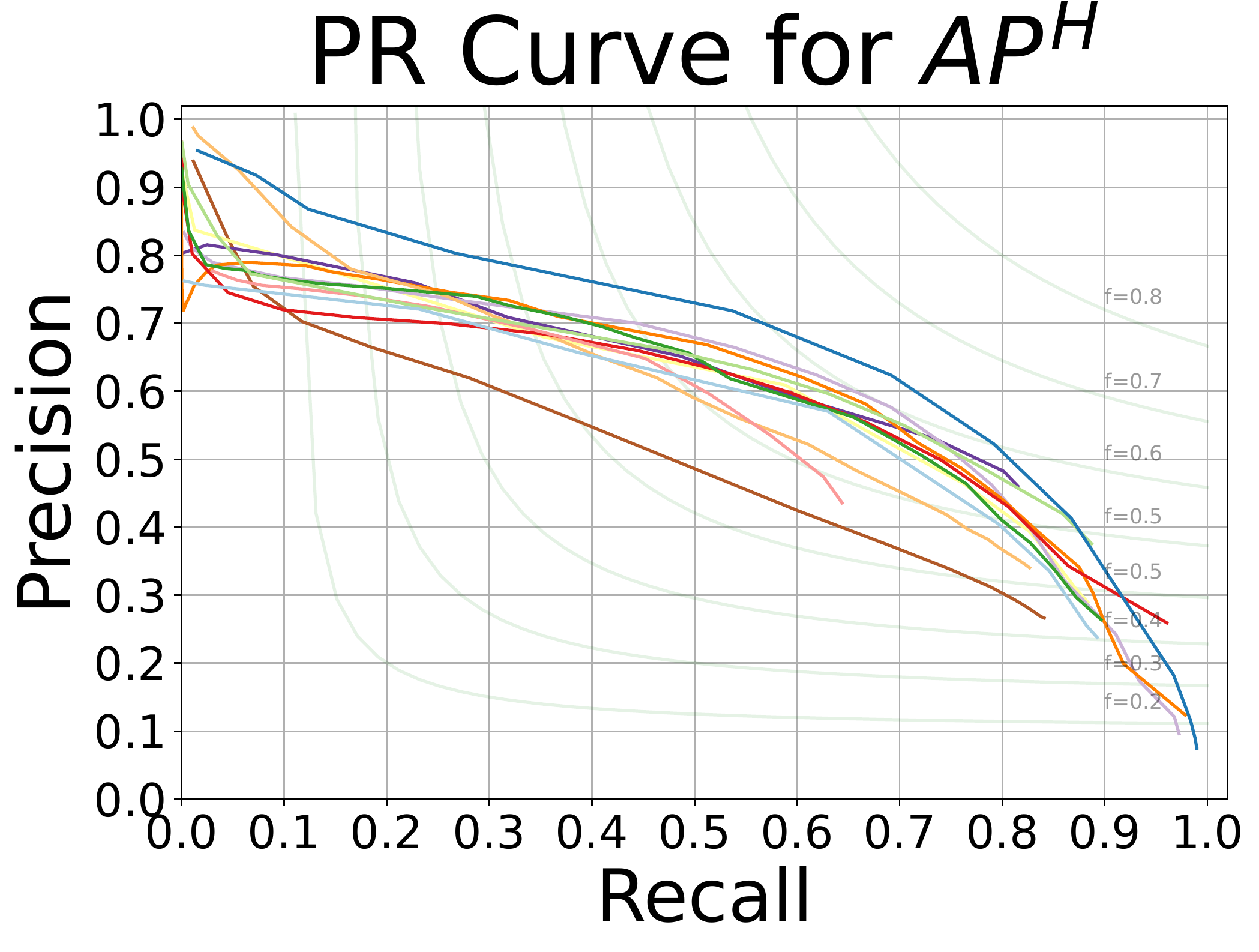}}
     \subfloat[fr3/cabinet]{\includegraphics[width=.245\linewidth]{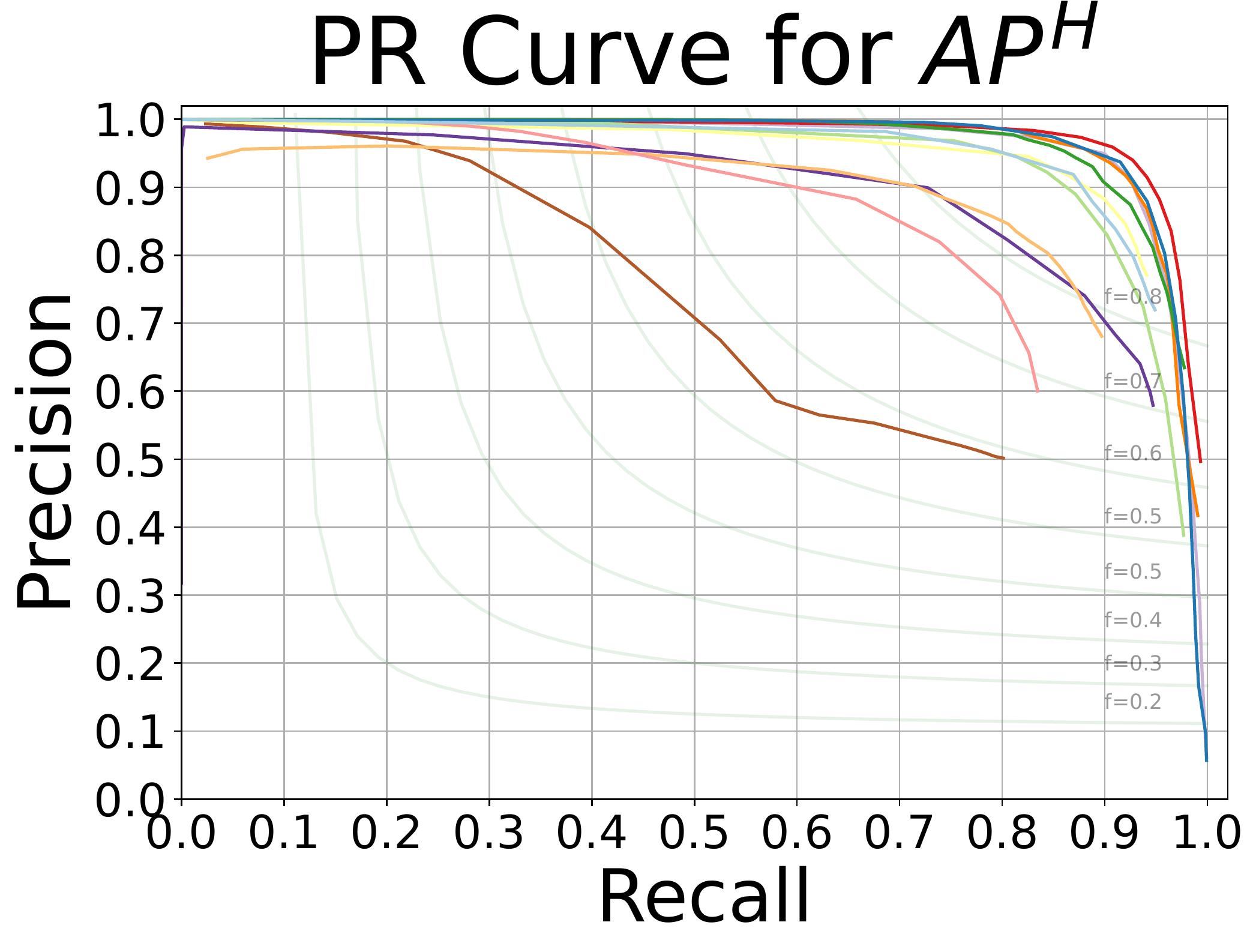}}\qquad
     \caption{Precision-Recall curves for $sAP^{10}$ and $AP^H$}
     \label{fig:pr_curves}
\end{figure*}

\begin{figure}[!htb]
     \centering
     \includegraphics[width=.24\linewidth]{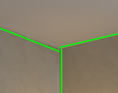}
     \includegraphics[width=.24\linewidth]{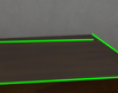}
     \includegraphics[width=.24\linewidth]{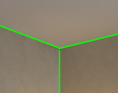}
     \includegraphics[width=.24\linewidth]{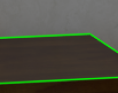}
     \caption{Detected lines displacement: F-Clip (two images on the left) and LSD (two images on the right).}
     \label{fig:line_displacement}
\end{figure}

The results of the detectors comparison are provided in~\cref{table:vectorized_scored,table:vectorized_unscored,table:heatmap_scored,table:heatmap_unscored,table:repeatability} and in~\cref{fig:pr_curves}. It can be seen that many detectors based on neural networks, such as HAWP, F-Clip, and LETR, have similar results and significantly outperform hand-crafted algorithms, such as LSD and ELSED, on vectorized and heatmap metrics. At the same time, hand-crafted algorithms show excellent results on repeatability metrics, probably because the line segments detected by neural networks are often displaced relative to the actual lines, as seen in~\cref{fig:line_displacement}, which leads to incorrect reprojection.

\begin{table}[!htb]
\centering
\caption{Comparison of detectors with a lines uncertainty measure: average heatmap precision ($AP^H$) and heatmap F-score ($F^H$).}
\label{table:heatmap_scored}
\huge
\resizebox{1\linewidth}{!}{
\begin{tabular}{|l||ll|ll|ll|ll|} 
\hline
\multirow{2}{*}{Algorithm} & \multicolumn{2}{c|}{lr kt2}   & \multicolumn{2}{c|}{of kt2}   & \multicolumn{2}{c|}{fr1/desk} & \multicolumn{2}{c|}{fr3/cabinet}  \\ 
\cline{2-9}
                           & { $AP^H$ }      & { $F^H$ }        & {$AP^H$ }        & { $F^H$ }         & { $AP^H$ }        & { $F^H$ }         & { $AP^H$ }        & { $F^H$ }             \\ 
\hline
AFM                        & 62.7          & 75.6          & 65.1          & 75.8          & 64.3          & 78.0          & 41.1          & 55.7              \\
ELSED                      & 57.1          & 68.4          & 52.8          & 68.0          & 80.5          & 82.5          & 52.2          & 56.1              \\
F-Clip                     & 86.5          & 83.0          & 87.8          & 81.0          & 94.8          & 92.5          & 59.4          & \uline{63.0}      \\
HAWP                       & \textbf{91.7} & \textbf{85.7} & \uline{89.1}  & \textbf{82.1} & 96.5          & 92.0          & \uline{59.6}  & 61.6              \\
HT-LCNN                    & 87.5          & 82.4          & 87.2          & 81.5          & 91.5          & 89.3          & 55.8          & 60.2              \\
L-CNN                      & 87.1          & 82.2          & 87.1          & \textbf{82.1} & 92.5          & 89.1          & 54.3          & 59.2              \\
LETR                       & 89.6          & 82.8          & \textbf{89.3} & \uline{81.7}  & \uline{96.8}  & \uline{93.0}  & \textbf{65.7} & \textbf{65.6}     \\
LSD                        & 58.0          & 65.1          & 45.3          & 64.3          & 61.0          & 61.7          & 44.3          & 49.3              \\
LSDNet                     & 77.6          & 77.8          & 74.2          & 80.1          & 87.3          & 82.0          & 54.9          & 61.0              \\
LSWMS                      & 32.7          & 57.3          & 37.6          & 58.0          & 39.2          & 62.8          & 24.1          & 46.7              \\
M-LSD                      & 82.7          & 77.8          & 79.8          & 76.3          & 92.9          & 88.6          & 57.4          & 62.1              \\
TP-LSD                     & \uline{90.2}  & \uline{85.1}  & 85.1          & 79.3          & \textbf{97.4} & \textbf{93.5} & 57.2          & 60.9              \\
ULSD                       & 88.7          & 84.7          & 86.2          & 79.4          & 95.3          & 91.0          & 56.6          & 60.1              \\
\hline
\end{tabular}
}
\end{table}
\begin{table}[!htb]
\centering
\caption{Comparison of detectors without a lines uncertainty measure: heatmap precision ($P^H$), recall ($R^H$), and F-score ($F^H$).}
\label{table:heatmap_unscored}
\huge
\resizebox{1\linewidth}{!}{
\begin{tabular}{|l||lll|lll|lll|lll|} 
\hline
\multirow{2}{*}{Algorithm} & \multicolumn{3}{c|}{lr kt2}                   & \multicolumn{3}{c|}{of kt2}                   & \multicolumn{3}{c|}{fr1/desk}                 & \multicolumn{3}{c|}{fr3/cabinet}               \\ 
\cline{2-13}
                           & $P^H$         & $R^H$         & $F^H$         & $P^H$         & $R^H$         & $F^H$         & $P^H$         & $R^H$         & $F^H$         & $P^H$         & $R^H$         & $F^H$          \\ 
\hline
EDLines                    & 52.5          & \textbf{86.4} & 65.3          & 61.6          & \textbf{78.2} & \uline{68.9}  & 26.5          & \textbf{91.6} & 41.1          & 57.9          & \textbf{93.5} & 71.5           \\
FLD                        & 53.6          & \uline{79.3}  & 64.0          & 60.9          & \uline{72.6}  & 66.2          & 28.1          & \uline{86.4}  & 42.4          & \uline{66.1}  & \uline{87.6}  & \uline{75.3}   \\
FSG                        & \uline{57.4}  & 78.8          & \uline{66.4}  & \uline{63.6}  & 67.7          & 65.6          & \uline{29.6}  & 83.1          & \uline{43.6}  & 53.0          & 87.2          & 65.9           \\
SOLD2                       & \textbf{74.5} & 66.6          & \textbf{70.3} & \textbf{81.1} & 67.5          & \textbf{73.7} & \textbf{62.9} & 36.2          & \textbf{45.9} & \textbf{87.0} & 66.9          & \textbf{75.6}  \\
\hline
\end{tabular}
}
\end{table}

\begin{table*}[!htb]
\centering
\caption{Repeatability metrics.}
\label{table:repeatability}
\resizebox{1\linewidth}{!}{
\begin{tabular}{|l||ll|ll|ll|ll|ll|ll|} 
\hline
\multirow{3}{*}{Algorithm} & \multicolumn{4}{c|}{lr kt2}                                               & \multicolumn{4}{c|}{of kt2}                                               & \multicolumn{4}{c|}{fr3/cabinet}                                           \\ 
\cline{2-13}
                           & \multicolumn{2}{c|}{$d_s$}          & \multicolumn{2}{c|}{$d_{\perp}$}    & \multicolumn{2}{c|}{$d_s$}          & \multicolumn{2}{c|}{$d_{\perp}$}    & \multicolumn{2}{c|}{$d_s$}          & \multicolumn{2}{c|}{$d_{\perp}$}     \\ 
\cline{2-13}
                           & Rep-5$\uparrow$ & Loc-5$\downarrow$ & Rep-5$\uparrow$ & Loc-5$\downarrow$ & Rep-5$\uparrow$ & Loc-5$\downarrow$ & Rep-5$\uparrow$ & Loc-5$\downarrow$ & Rep-5$\uparrow$ & Loc-5$\downarrow$ & Rep-5$\uparrow$ & Loc-5$\downarrow$  \\ 
\hline
AFM@0.1                    & 0.21            & 2.48              & 0.44            & 1.79              & 0.19            & 2.51              & 0.40            & 1.66              & 0.08            & 2.49              & 0.24            & 2.51               \\
ELSED@0                    & 0.33            & 2.56              & 0.51            & 1.43              & 0.36            & 2.44              & \uline{0.52}    & \textbf{1.28}     & 0.18            & 2.91              & 0.37            & 1.80               \\
F-Clip@0.4                 & 0.34            & 2.24              & 0.53            & 1.57              & 0.34            & \uline{2.06}      & 0.47            & 1.41              & \textbf{0.27}   & 2.54              & 0.37            & 1.91               \\
HAWP@0.8                   & 0.35            & \textbf{2.16}     & \textbf{0.55}   & 1.52              & 0.35            & 2.11              & 0.47            & 1.41              & \textbf{0.27}   & 2.41              & 0.35            & 1.82               \\
HT-LCNN@0.98               & 0.32            & \uline{2.22}      & 0.51            & 1.55              & 0.31            & \textbf{2.01}     & 0.45            & 1.36              & 0.23            & \textbf{2.20}     & 0.30            & 1.75               \\
L-CNN@0.98                 & 0.33            & 2.27              & 0.53            & 1.55              & 0.31            & 2.07              & 0.45            & 1.40              & 0.25            & 2.43              & 0.33            & 1.86               \\
LETR@0.7                   & 0.33            & 2.45              & 0.53            & 1.67              & 0.35            & 2.44              & 0.48            & 1.66              & 0.30            & 2.59              & 0.39            & 1.92               \\
LSD@0                      & \textbf{0.43}   & 2.56              & 0.52            & \textbf{1.40}     & \textbf{0.45}   & 2.41              & 0.51            & 1.31              & 0.21            & 2.89              & 0.35            & \uline{1.73}       \\
LSDNet@0.8                 & 0.30            & 2.44              & 0.50            & 1.58              & 0.30            & 2.33              & 0.45            & 1.41              & 0.20            & 2.53              & 0.34            & 1.90               \\
LSWMS@0                    & 0.28            & 3.04              & 0.41            & 1.71              & 0.20            & 3.19              & 0.31            & 2.09              & 0.16            & 3.28              & 0.23            & 2.16               \\
M-LSD@0.4                  & 0.33            & 2.42              & 0.50            & 1.57              & 0.35            & 2.25              & 0.47            & 1.51              & 0.24            & 2.59              & 0.32            & 1.89               \\
TP-LSD@0.3                 & 0.32            & 2.30              & 0.52            & 1.53              & 0.36            & 2.21              & 0.49            & 1.43              & \textbf{0.27}   & 2.51              & 0.37            & 1.89               \\
ULSD@0.9                   & 0.34            & 2.25              & \uline{0.54}    & 1.54              & 0.32            & 2.09              & 0.46            & 1.42              & \uline{0.26}    & 2.48              & 0.35            & 1.85               \\
EDLines                    & \uline{0.40}    & 2.60              & 0.51            & \textbf{1.40}     & \uline{0.43}    & 2.52              & 0.51            & 1.31              & \uline{0.26}    & 3.00              & \textbf{0.44}   & 1.76               \\
FLD                        & 0.39            & 2.70              & 0.51            & \uline{1.42}      & \uline{0.43}    & 2.58              & 0.50            & \uline{1.29}      & 0.21            & 2.97              & \uline{0.41}    & 1.83               \\
FSG                        & 0.25            & 2.53              & 0.52            & 1.49              & 0.27            & 2.48              & \textbf{0.53}   & 1.44              & 0.14            & 2.66              & 0.34            & \textbf{1.72}      \\
SOLD2                      & 0.38            & 2.59              & 0.47            & 1.51              & 0.42            & 2.35              & 0.47            & 1.33              & 0.15            & \uline{2.39}      & 0.28            & 1.78               \\
\hline
\end{tabular}
}
\end{table*}

\begin{table}[!htb]
\centering
\caption{Associators comparison: precision ($P$), recall ($R$), F-score ($F$), and FPS}
\label{table:association}
\huge
\resizebox{1\linewidth}{!}{
\begin{tabular}{|l||lll|l|lll|l|lll|l|lll|l|} 
\hline
\multirow{2}{*}{Algorithm} & \multicolumn{4}{c|}{lr kt2}                                 & \multicolumn{4}{c|}{of kt2}                                 & \multicolumn{4}{c|}{fr1/desk}                                & \multicolumn{4}{c|}{fr3/cabinet}                              \\ 
\cline{2-17}
                           & P             & R             & F             & FPS         & P             & R             & F             & FPS         & P             & R             & F             & FPS          & P             & R             & F             & FPS           \\ 
\hline
LBD                        & 84.8          & \uline{69.6}  & \uline{76.4}  & \textbf{94} & 86.0          & \uline{73.7}  & \uline{79.4}  & \textbf{80} & 36.6          & \uline{25.2}  & \uline{29.8}  & \textbf{109} & 83.3          & 50.9          & 63.2          & \textbf{195}  \\
DLD                        & 81.8          & 64.4          & 72.1          & \uline{30}  & 78.3          & 64.7          & 70.8          & \uline{16}  & 23.9          & 16.3          & 19.4          & \uline{23}   & 73.6          & 49.1          & 58.9          & \uline{74}    \\
WLD                        & 82.5          & 66.7          & 73.8          & 4           & 80.3          & 61.8          & 69.8          & 2           & 35.5          & 22.7          & 27.7          & 3            & 86.7          & \uline{63.8}  & \uline{73.5}  & 13            \\
LineTR                     & \textbf{96.8} & 56.7          & 71.5          & 14          & \textbf{95.7} & 61.6          & 75.0          & 12          & \textbf{69.5} & 6.8           & 12.5          & 14           & \textbf{96.9} & 41.2          & 57.8          & 19            \\
SOLD2                      & \uline{88.1}  & \textbf{80.8} & \textbf{84.3} & 2           & \uline{89.2}  & \textbf{82.2} & \textbf{85.5} & 1           & \uline{42.3}  & \textbf{32.4} & \textbf{36.7} & 2            & \uline{94.2}  & \textbf{84.6} & \textbf{89.1} & 4             \\
\hline
\end{tabular}
}
\end{table}
\begin{table}[!htb]
\centering
\caption{Pose error estimation: median translation error and rotation error; a dash means that the pose could not be calculated in more than half of the cases.}
\label{table:pose_error}
\huge
\resizebox{1\linewidth}{!}{
\begin{tabular}{|l|l||ll|ll|ll|} 
\hline
\multirow{2}{*}{Detector}   & \multirow{2}{*}{Associator} & \multicolumn{2}{c|}{lr kt2}       & \multicolumn{2}{c|}{of kt2}       & \multicolumn{2}{c|}{fr3/cabinet}   \\ 
\cline{3-8}
                            &                             & $\epsilon_{trans}$ & $\epsilon_{rot}$ & $\epsilon_{trans}$ & $\epsilon_{rot}$ & $\epsilon_{trans}$ & $\epsilon_{rot}$  \\ 
\hline
\multirow{5}{*}{LSD}        & LBD                         & 0.452            & 7.888          & 0.446            & 6.986          & 1.074            & 28.052          \\
                            & DLD                         & 0.206            & 3.589          & 0.315            & 4.585          & 1.067            & 32.883          \\
                            & WLD                         & 0.383            & 7.259          & 0.765            & 13.199         & 0.818            & 22.637          \\
                            & LineTR                      & \textbf{0.025}   & \textbf{0.890} & \textbf{0.026}   & \textbf{0.683} & 0.620            & 17.384          \\
                            & SOLD2                       & 0.144            & 2.902          & 0.099            & 1.917          & \textbf{0.162}   & \textbf{5.319}  \\ 
\cline{1-2}
\multirow{5}{*}{F-Clip@0.4} & LBD                         & 0.790            & 12.950         & 0.948            & 14.130         & —                & —               \\
                            & DLD                         & 0.306            & 5.321          & 0.944            & 13.025         & 67.799           & 153.450         \\
                            & WLD                         & 0.594            & 8.396          & 1.599            & 25.394         & 3.114            & 71.209          \\
                            & LineTR                      & 0.061            & 1.579          & \uline{0.074}    & \uline{1.517}  & —                & —               \\
                            & SOLD2                       & 0.095            & 2.164          & 0.295            & 3.968          & \uline{0.431}    & \uline{12.772}  \\ 
\cline{1-2}
\multirow{5}{*}{HAWP@0.7}   & LBD                         & 1.993            & 34.954         & 2.050            & 32.943         & —                & —               \\
                            & DLD                         & 1.462            & 23.861         & 1.904            & 27.084         & —                & —               \\
                            & WLD                         & 1.533            & 25.016         & 2.425            & 44.257         & —                & —               \\
                            & LineTR                      & \uline{0.058}    & 1.602          & 0.079            & 1.521          & —                & —               \\
                            & SOLD2                       & \uline{0.058}    & \uline{1.383}  & 0.356            & 4.202          & 0.517            & 16.713          \\
\hline
\end{tabular}
}
\end{table}

\subsection{Line Association}
For association comparison, we use our line annotations and the following metrics: precision, recall, and F-score. Comparison of results are presented in~\cref{table:association}. It is noticeable that LineTR has the highest precision among the presented algorithms but relatively low recall. SOLD2, in turn, outperforms the rest of the algorithms regarding recall and F-score. It is also worth clarifying that LBD also shows quite good results.

\subsection{Pose estimation}
Relative pose estimations, as described in Sec.~\ref{sec:pose_est}, are derived from the line matches obtained using each association and three popular detectors: LSD, F-Clip, and HAWP. The ground truth poses are obtained by transforming the absolute poses provided by the authors of the original datasets. Then estimated pose is compared with ground truth pose and classical Relative Pose Error (RPE) is calculated for rotation~($\epsilon_{rot}$) and translation~($\epsilon_{trans}$).

The results are presented in \cref{table:pose_error}. It can be seen that the LSD + LineTR combination shows superior results on all datasets except \texttt{fr3/cabinet}. Also, the results of association SOLD2 and LineTR with all three detectors are superior to the rest of the considered detector-association combinations. The reason why LSD produces good results, probably, as in the case of repeatability metrics, is better line reprojection.

\subsection{Performance}
In addition to evaluating the quality of detectors and associations, we evaluate their performance. We measure the speed of work on all datasets. For all neural networks, we set the batch size to 1.
It is important to note that we compare all the algorithms using python code so that the results may differ from the original implementations. It is also worth considering that we run LSDNet on CPU since the authors provided such a version of the algorithm.
The results obtained are presented in~\cref{table:vectorized_scored,table:vectorized_unscored,table:association}. It is easy to see that handcrafted detectors, for example, EDLines and FLD, significantly outperform neural network algorithms regarding speed. However, it should be noted that there are quite fast algorithms among neural network detectors, for example, M-LSD. Neural network association, in turn, significantly loses to LBD in terms of speed.

\section{Conclusion}
In this paper, we have presented a complete and exhaustive comparison for using visual lines in a SLAM front-end. To do so, we have labeled popular SLAM sequences from ICL NUIM and TUM RGB-D datasets. In particular, we have provided 17 line detection algorithms, 5 line associations methods and the resultant pose error or aligning a pair of frames with several combinations of detector-association. These results are the solid ground from which to benchmark the visual SLAM front-end involving lines, in all its stages: from detection to its effect in pose error.

\begin{spacing}{0.8}
\bibliographystyle{IEEEtran} 
\bibliography{bibliography}
\end{spacing}

\end{document}